\documentclass[lettersize,journal]{IEEEtran}
\usepackage{amsmath,amsfonts}
\usepackage{algorithmic}
\usepackage{algorithm}
\usepackage{array}
\usepackage[caption=false,font=normalsize,labelfont=sf,textfont=sf]{subfig}
\usepackage{textcomp}
\usepackage{stfloats}
\usepackage{url}
\usepackage{verbatim}
\usepackage{graphicx}
\usepackage{cite}
\usepackage{color}
\usepackage{multirow}

\usepackage{xcolor}

\let\textquotedbl="

\begin{document}

\title{Zero-shot Cross-lingual Stance Detection via Adversarial Language Adaptation}

\author{
\IEEEauthorblockN{Bharathi A\IEEEauthorrefmark{1},
and Arkaitz Zubiaga\IEEEauthorrefmark{2}}\\
\IEEEauthorblockA{\IEEEauthorrefmark{1}PSG College of Technology, India}
\IEEEauthorblockA{\IEEEauthorrefmark{2}Queen Mary University of London, UK}
\thanks{Manuscript received April 19, 2021; revised August 16, 2021.}
}



\maketitle

\begin{abstract}
 Stance detection has been widely studied as the task of determining if a social media post is positive, negative or neutral towards a specific issue, such as support towards vaccines. Research in stance detection has however often been limited to a single language and, where more than one language has been studied, research has focused on few-shot settings, overlooking the challenges of developing a zero-shot cross-lingual stance detection model. This paper makes the first such effort by introducing a novel approach to zero-shot cross-lingual stance detection, Multilingual Translation-Augmented BERT (MTAB), aiming to enhance the performance of a cross-lingual classifier in the absence of explicit training data for target languages. Our technique employs translation augmentation to improve zero-shot performance and pairs it with adversarial learning to further boost model efficacy. Through experiments on datasets labeled for stance towards vaccines in four languages --English, German, French, Italian--, we demonstrate the effectiveness of our proposed approach, showcasing improved results in comparison to a strong baseline model as well as ablated versions of our model. Our experiments demonstrate the effectiveness of model components, not least the translation-augmented data as well as the adversarial learning component, to the improved performance of the model. We have made our source code\footnote{\url{https://github.com/Bharathi-A-7/MTAB-cross-lingual-vaccine-stance-detection}} accessible on GitHub.
\end{abstract}


\section{Introduction}

\IEEEPARstart{S}{tance} detection is a widely studied Natural Language Processing (NLP)
task that consists in determining the viewpoint expressed in a text
towards a particular target \cite{kuccuk2020stance,aldayel2021stance}.
By determining the viewpoints of a large collection of posts collected
through online sources such as social media, one can then get a sense
of the public opinion towards the target in question \cite{cao2022stance,almadan2023stance}.
Given a text as input, a stance detection model then determines if
it conveys a supporting (positive), opposing (negative) or neutral viewpoint towards a target.
Recent research has therefore looked into improving stance detection
models that make competitive predictions across all three classes, in domains such as politics \cite{liu2022politics} and healthcare \cite{conforti2020will}. 

In line with recent progress in NLP research \cite{min2023recent,zubiaga2024natural}, state-of-the-art stance
detection models embrace methods based on large language models
such as Transformers \cite{karande2021stance,gunhal2022stance,zhao2023c}. However, where stance detection is of global
interest, research has generally studied the applicability of models
to a single language, predominantly English, and generalizability
to other languages has been understudied. Stance detection models,
primarily trained on a single language, face challenges when applied
to multilingual scenarios \cite{lai2020multilingual}. The paucity of labeled data for each
language further complicates this task, demanding innovative strategies
to navigate linguistic diversity effectively, particularly in the
case of low-resource languages where labeled data is scarce or unavailable.
Where limited prior research has studied cross-lingual methods to
stance detection \cite{Hardalov_etal_2022_Few-Shot,Mohtarami_etal_2019_Contrastive},
there has been a lack of endeavors into developing zero-shot cross-lingual
approaches that can be applied when there is no labeled data available
in the target language. This is however critical for lesser-resourced
languages where labeling relevant datasets is often unaffordable \cite{cao2019low,ding2020daga,hedderich2021survey}. 

In this work, we present the first such method into zero-shot cross-lingual
stance detection, moving away from previous studies on few-shot cross-lingual
stance detection which require some amount of labeled data from the target language
to be available during training. To achieve this, we introduce Multilingual
Translation-Augmented BERT (MTAB), a novel
architecture that integrates translation augmentation, multilingual
encoder and adversarial language adaptation through a transformer
model for effective stance detection applicable to languages for which
no labeled data has been seen during training. To test the effectiveness
of MTAB, we perform experiments on datasets in four different
languages (i.e. English, French, German and Italian) for the same
domain, namely stance detection around vaccine hesitancy. Our experiments
show that this combination consistently outperforms cross-lingual transfer in models
that only use adversarial learning as a technique to align embeddings
to the target languages. Our research establishes a benchmark opening up a novel research direction into zero-shot cross-lingual stance detection.

Our work makes the following novel contributions:

\begin{itemize}
 \item We introduce MTAB, a novel approach to zero-shot cross-lingual
stance detection that uses a translation-augmented training dataset
and an adversarial learning model to adapt to unseen languages with
no labelled examples. We believe that our approach is not only applicable
to this particular task, but can also be extended to other cross-lingual
text classification problems.

 \item To the best of our knowledge, ours is the first work addressing
zero-shot cross-lingual stance detection. Additionally, it is the
first study to apply an adversarial learning approach to cross-lingual
stance detection. 

 \item Cross-lingualism remains a less-explored topic in stance detection,
and our research aims to bridge that gap. As one of the first studies
in this area, we delve into the potential of standard Natural Language
Processing (NLP) and transfer learning techniques to tackle cross-lingual
stance detection, contributing to the evolving understanding of this
domain.
\end{itemize}

The remainder of the paper is organized as follows. Next, in Section \ref{sec:related-work}, we discuss prior research relevant to ours and we highlight our key novel contributions on top of those. We then describe the datasets we use in our study in Section \ref{ref:datasets}. We introduce and describe our newly proposed model, MTAB, in Section \ref{sec:mtab}, followed by details of our experimental settings in Section \ref{sec:experiments}. We show and analyze results of our experiments in Section \ref{sec:results}, subsequently concluding the paper in Section \ref{sec:conclusion}.

\section{Related Work}
\label{sec:related-work}

The ability to accurately classify stances on unseen or unfamiliar
topics or languages is of paramount importance in real-world applications,
where issues and discussions are constantly evolving \cite{alkhalifa2022capturing,alkhalifa2023building}.
Traditional stance detection models often struggle to generalize effectively
across different topics, domains, and languages, limiting their practical
utility \cite{Reuver_etal_2021_IsStance,schiller2021stance,ng2022my}. To address this limitation,
recent research has focused on developing models capable of generalizable
stance detection. These efforts have involved exploring transfer learning,
domain adaptation and other cross-lingual approaches to enhance the
model's ability to handle unseen or unfamiliar topics
and languages. 

\subsection{Cross-topic and cross-target stance detection}

Cross-topic stance detection aims to detect stance across topics using
knowledge transfer. This approach leverages large stance datasets
from source topics to improve performance on smaller target topic
datasets. However, most existing work in cross-topic stance detection
assumes some labelled data in the target topic is available for training.
Zero-shot cross-topic stance detection aims to detect stance in an
entirely novel target topic with no labelled training data, instead
leveraging source topic data and topic-independent features. Though
still relatively understudied, several approaches have been proposed
for zero-shot cross-topic stance detection, including training on
multi-target datasets, topic-invariant representation learning and
unsupervised domain adaptation methods which can also be applied to
scenarios involving cross-lingual and cross-domain problems. 

Zarrella and Marsh (2016) proposed a system for the SemEval-2016 competition
on stance detection in tweets that uses transfer learning from unlabeled
datasets and an auxiliary hashtag prediction task to learn sentence
embeddings. Their approach, which achieved the top score in the task,
uses a recurrent neural network (RNN) initialized with features learned
via distant supervision on two large unlabeled datasets, and then
fine-tuned on a small dataset of labeled tweets \cite{Zarrella_Marsh_2016_MITRE}.
For the same task, Augenstein et al. (2016) train a Any-Target stance
classifier using a bag-of-words autoencoder that learns feature representations
independent of any target \cite{Augenstein_etal_2016_USFD}. Xue et
al. (2018) and Wei and Mao. (2019) proposed two different approaches
for cross-target stance classification, both of which were able to
achieve state-of-the-art results on the SemEval-2016 Task 6 test set.
Xue et al. used a self-attention network to extract target-independent
information for model generalization \cite{Xu_etal_2018_cross_target},
while Wei and Mao (2019) introduced a topic modeling approach that
can leverage shared latent topics between two targets as transferable
knowledge to generalize across topics \cite{Wei_Mao_2019_Modeling}.
Building upon these advancements, several other notable contributions
have been made in achieving state-of-the-art results in cross-topic
stance detection, mainly within the context of the SemEval-2016 Task
6. Wang et al. (2020) employed adversarial domain generalization to
learn target-invariant text representations \cite{Wang_etal_2020_unseen}.
In addition, Hardalov et al. (2021) combined domain adversarial learning
with label-adaptive learning to learn input representation specific
to the stance target domain, and handle unseen domains using embedding
similarities \cite{Hardalov_etal_2021_Crossdomain}. A more recent
work by Ji et al. (2022) establishes a new state-of-the-art for the
task by incorporating a model based on meta-learning, which is also
trained on multiple targets considering the similarities between targets
\cite{Ji_etal_2022_Cross_Target}. More recently, Jamadi Khiabani
and Zubiaga (2023) introduced a model that leverages features derived
from the social network, in addition to the textual content of posts,
to improve a model for cross-target stance detection \cite{khiabani2023few},
which they showed could be further used to enhance a language model
for stance detection \cite{jamadi2024socialpet}. 

Recent approaches in stance detection for unseen targets, or more
formally zero-shot stance detection have shown promising results.
One of the first works in this area by Allaway and McKeown (2020),
introduced a new dataset VAST for zero shot stance detection and developed
a model that learns generalized topic representations to pre- dict
stance labels based on similarities between given topics and the topics
present in training set \cite{Allaway_McKeown_2020_Zero_Shot}. Allaway
et al. (2021) also proposed a new TOpic ADversarial Network (TOAD)
that conditions document representations on topics and further learns
domain-invariant features using adversarial training for zero-shot
stance detection \cite{Allaway_etal_2021_Adversarial}. Liang et al.
(2022) proposed a novel approach for ZSSD that frames the distinction
of target-invariant and target-specific stance features as a pre-text
task to better learn transferable stance features \cite{Liang_etal_2022_Zero_Shot}.
In the case of Twitter, an unsupervised zero-shot stance detection
framework Tweet2Stance, proposed by Gambini et al. (2022) uses content
analysis of users' Twitter timelines with an NLI based
zero-shot classifier to infer their stance towards a political-social
statement \cite{Gambini_etal_2022_Tweets2Stance}. 

\subsection{Cross-lingual stance detection}

Current approaches to stance detection are well-studied in English
but have received less attention in other languages and cross-lingual
settings. A novel approach to improve cross-lingual stance detection
introduced by Mohtarami et al. (2019) uses memory networks along with
a contrastive language adaptation component to effectively align source
and target language data with the same labels and separate the ones
with different labels \cite{Mohtarami_etal_2019_Contrastive}. To
facilitate research in multilingual stance detection, Zotova et al.
(2020) introduced a new Twitter dataset for the Catalan and Spanish
languages annotated with Stance towards the independence of Catalonia
\cite{Zotova_etal_2020_Multilingual}. X-stance is another multilingual
stance detection dataset introduced by Vamvas and Sennrich (2020)
that includes text in German, French and Italian, covering a wide
range of topics in Swiss politics \cite{Vamvas_Sennrich_2020_Xstance}.
A more recent work by Hardalov et al. (2022) presents a comprehensive
study of cross-lingual stance detection, covering 15 datasets in 12
languages that demonstrates the effectiveness of pattern-based training
approaches and pre-training using labeled instances from a sentiment
model \cite{Hardalov_etal_2022_Few-Shot}. 

Existing research on cross-lingual stance detection has however been
limited to few-shot settings, where a small portion of labeled data
is available for the target language. In our work, we are interested
in going further by enabling zero-shot cross-lingual stance detection
where there is no labeled data at all for the target language. With
this objective in mind, we propose our novel model MTAB, which
leverages adversarial learning through translation-augmented training
data for zero-shot cross-lingual stance detection. 

\subsection{Vaccine stance detection}

The task of vaccine stance detection is still relatively new, but
there has been significant progress in recent years, given the rise
of vaccine misinformation on social media. Twitter is a rich source
of data for vaccine discourse, and many researchers have used Twitter
data to develop mono-lingual and cross-lingual vaccine stance detection
models. The method introduced by Bechini et al. \cite{Bechini_etal_2021_Italy_Vaccination}
particularly demonstrates the importance of incorporating user related
information in the analysis of public opinion on Twitter and combine
that with text classification techniques to asses opinions on vaccination
topic in Italy. In the month following the announcement of the first
COVID-19 vaccine, Cotfas et al. (2021) provided insights into the
dynamics of opinions on COVID-19 vaccination based on tweets and compared
classical machine learning and deep learning algorithms for stance
detection \cite{Cotfas_etal_2021_Longest_Month}. In a related study
conducted on the same period, the authors examined vaccine hesitancy
discourse on Twitter using several text analysis techniques such as
LDA, Hashtag and N-gram analysis \cite{Cotfas_etal_2021_Month_Following}.
Addressing this challenge in low-resource languages, Küçük and Ar\i c\i{}
(2022) introduced the first dataset for Turkish tweets that is annotated
with stance and sentiment labels towards COVID-19 vaccination \cite{Kucuk_Arici_2022_Turkish}.
In addition, the first manually annotated Arabic tweet dataset related
to the COVID-19 vaccination campaign (ArCovidVac) proposed by Mubarak
et al. (2022) consists of stance annotated tweets from different Arab
countries and delivers key insights into the discussions and opinions
related to the COVID-19 vaccination campaign on social media in the
Arab region \cite{Mubarak_etal_2022_ArCovidVac}. 

The domain of vaccine hesitancy presents an ideal scenario for our
research, as it is an issue of global interest, which is discussed
and opinionated across languages in the world, which has in turn led
to the creation of labeled datasets in different languages. In our
case, we use datasets in four languages, namely English, French, German
and Italian, to enable our zero-shot cross-lingual stance detection study. 

\section{Datasets}
\label{ref:datasets}

In this section, we provide a detailed description of the datasets
used for training and testing our model, along with the associated
pre-processing techniques. Specifically, we use English data only
for training, and then use three different test sets in three different
languages for testing, i.e. French, German and Italian. The statistics
of our datasets for both training and testing are provided in Table \ref{tab:datasets}.

\subsection{Training Datasets}

We experiment in the scenario where a model is only trained on a resource-rich language, English, to then test it on other languages. As such, our training data only contains data in the English language, while test data will be written in other languages. We therefore set up the scenario where the target languages found in the test sets are not seen during training.

To achieve this objective and create our English-only training data, we combined data from three different sources
of English tweets on COVID-19 vaccines belonging to different timelines
during the period of 2020 and 2021 to create a more diverse and representative
sample. We aggregate the following three datasets:

\begin{enumerate}
 \item \textbf{Vaccine Attitude Dataset (VAD) by Zhu et al. (2021):} The
VAD dataset \cite{Zhu_etal_2022_Disentagled}, originally introduced
by Zhu et al. (2021), consists of an extensive collection of 1.9 million
English tweets obtained between February 7th and April 3rd, 2021.
The authors annotated a random subset of 2800 tweets from this corpus
with stance labels and aspect characterizations. We use this smaller, labeled
subset as part of our training data. 

 \item \textbf{COVID-19 Vaccination Tweets dataset by Almadan et al. (2022):}
The work of Almadan et al. (2022) presented a systematic methodology
for annotating tweets with respect to their stance toward COVID-19
vaccination. This approach included the development of a comprehensive
codebook for stance annotation and the use of keyword and hashtag
based sampling techniques for Twitter data. Leveraging this coding
framework, the authors released a stance annotated dataset for research
purposes \cite{Almadan_etal_2022_Will}, which we also incorporate into our training dataset. 

 \item \textbf{COVID-19 Vaccine Stance Dataset by Cotfas et al. (2021):}
The authors Cotfas et al. (2021) collected 2,349,659 tweets related
to COVID-19 vaccination during the period of November 9th to December
8th, 2020 using a set of predefined keywords via the Twitter API. From this collection,
they curated a refined and balanced stance annotated dataset comprising
3,249 tweets, which was made publicly accessible \cite{Cotfas_etal_2021_Longest_Month}. We incorporate this smaller, labeled subset of their dataset into our training data.
\end{enumerate}

After aggregating the three labeled data sources, our final training data contains 4493 tweets, of which 2276 are labeled positive, 1141 negative, and 1076 neutral. See Table \ref{tab:Training-data} for more details on the training data, including the final distribution of labels.

\subsection{Test Data}

For testing our model on non-English data, we use data pertaining to three languages --French, German and Italian-- from the VaccinEU \cite{Giovanni_etal_2022_vaccin}
dataset as introduced by Giovanni et al. (2022). To study the impact
of online conversations on COVID-19 vaccines, the researchers used
a list of vaccine related keywords and collected a large dataset of
over 70 million tweets in three different languages, namely French, German,
and Italian, from November 1\textsuperscript{st}, 2020 to November
15\textsuperscript{th}, 2021. From this large-scale dataset, the authors labeled small samples for each language, providing four labels —Positive, Negative, Neutral, and Out of Context. However, for our experiments, we only focus on the Positive, Negative, and Neutral classes, as detailed in Table \ref{tab:Test-data}. We can observe that labels are unevenly distributed across different languages, which is itself an additional challenge for our research, but one can realistically expect that these variations across languages will naturally happen, not least because of cultural differences affecting level of support towards vaccines. Given the importance of label distributions in our cross-lingual research, we delve into performance analysis across labels in our analysis of results, in addition to looking at general performance scores aggregated for all labels.

\subsection{Data Pre-processing}

Due to Twitter's terms of service and privacy concerns,
these data were made available only using Tweet IDs, and hence, we
first used an additional application called Hydrator\footnote{\url{https://github.com/DocNow/hydrator}}
to retrieve the actual tweet content associated with the IDs. Due
to the age and controversial nature of the tweets, some of them were
lost during retrieval. This however happens generally with Twitter datasets, as research has shown that a percentage of tweets inevitably tends to disappear \cite{zubiaga2018longitudinal}, and hence we conduct our research with the tweets available at the time of our collection, which are the statistics we report in Table \ref{tab:datasets}.

In the first step of data pre-processing, we standardize the stance
labels of the three datasets that were combined for model training.
This ensures the diverse labeling conventions present in those datasets
are harmonized enabling consistent and integrated experimentation.

Once the data were ready, and to prepare the tweets for further analysis and modeling, we carried
out a series of textual pre-processing steps, as follows:

\begin{itemize}
 \item \textbf{Removal of URLs, Emojis, Mentions, Smileys, and Numbers:}
In the interest of focusing on the textual content of the tweets, we used the ``tweet-preprocessor''\footnote{\url{https://pypi.org/project/tweet-preprocessor/}} library
to eliminate URLs, emojis, user mentions, smileys, and numbers from the
tweet text. 

 \item \textbf{Handling Retweets:} Retweets often contain a standard
``RT @'' prefix that can affect the consistency
of the text by adding noise to it which provides no information about the stance of a tweet. We removed this prefix to standardize the format of retweets
while preserving the essence of their content. 

 \item \textbf{Handling Hashtags:} To maintain the relevance of hashtags
as keywords, we opted to remove only the `\#'
symbol, retaining the keyword itself. This approach ensured that the
hashtags contributed to the context of the tweet without introducing
any extraneous characters, while we retain the content of the hashtag whenever this forms part of the sentence, which we did not want to break. 
\end{itemize}

The resulting texts, after the three preprocessing steps above, are then fed to our models for training or testing.

\begin{table*}
\begin{centering}
\subfloat[Statistics of the tr\label{tab:Training-data}aining data]{\centering{}%
\begin{tabular}{|c|c|c|c|c|}
\hline 
\multicolumn{5}{|c|}{Training Set} \tabularnewline
\hline 
\hline 
\multirow{2}{*}{Data Source} & \multicolumn{3}{c|}{Label Distribution} & \multirow{2}{*}{Total \# of Tweets}\tabularnewline
\cline{2-4} 
 & Positive & Negative & Neutral & \tabularnewline
\hline 
Vaccine Attitude Dataset (VAD) by Zhu et al (2021) & 1367 & 517 & 162 & 2046\tabularnewline
\hline 
Almadan et al (2022) & 30 & 12 & 13 & 55\tabularnewline
\hline 
Cotfas et al (2021) & 879 & 612 & 901 & 2392\tabularnewline
\hline 
\hline
\textbf{Aggregated training data} & \textbf{2276} & \textbf{1141} & \textbf{1076} & \textbf{4493}\tabularnewline
\hline
\end{tabular}} 
\par\end{centering}
\begin{centering}
\subfloat[Statistics of the testing data\label{tab:Test-data}]{\centering{}%
\begin{tabular}{|c|c|c|c|c|}
\hline 
\multicolumn{5}{|c|}{Testing Set - VaccinEU} \tabularnewline
\hline 
\hline 
\multirow{2}{*}{Target Language} & \multicolumn{3}{c|}{Label Distribution} & \multirow{2}{*}{Total \# of Tweets}\tabularnewline
\cline{2-4}
 & Positive & Negative & Neutral & \tabularnewline
\hline 
French & 419 & 135 & 279 & 833\tabularnewline
\hline 
German & 547 & 108 & 169 & 824\tabularnewline
\hline 
Italian & 314 & 151 & 458 & 923\tabularnewline
\hline 
\end{tabular}}
\par\end{centering}
\centering{}\caption{Dataset statistics. We use the aggregated training data for training our models, which contains tweets only in English. We test our models on tweets in unseen languages individually, i.e. French, German and Italian.}
\label{tab:datasets}
\end{table*}

\section{MTAB: Model Architecture}
\label{sec:mtab}

\begin{figure*}
\begin{centering}
\includegraphics[width=\textwidth]{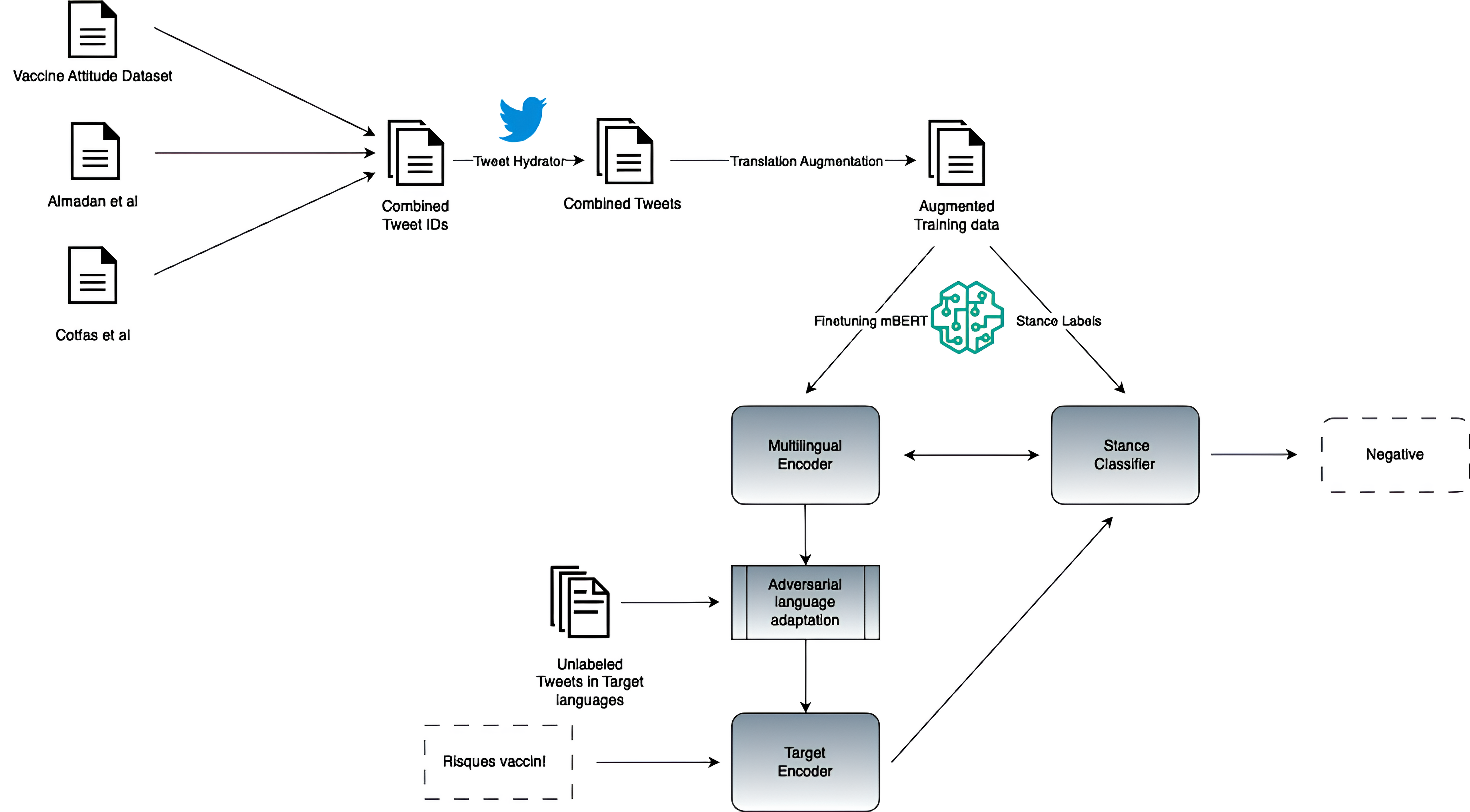}
\par\end{centering}
\caption{\label{fig:Proposed-Architecture}Architecture of our proposed MTAB model.}

\end{figure*}

Here we introduce the architecture of our zero-shot cross-lingual
stance detection model, Multilingual Translation-Augmented BERT (MTAB), which we design to categorize public opinions on
vaccines across diverse languages, specifically targeting French,
German, and Italian. Our architecture employs two levels of data augmentation
combined with training a neural stance classifier on multilingual
contextual embeddings that are adapted to the target languages via
adversarial training. The full architecture is presented in Figure
\ref{fig:Proposed-Architecture}, and we describe each of the three components
next: (i) translation augmentation, (ii) multilingual encoder and stance classifier, 
and (iii) adversarial language adaptation.

\subsection{Translation Augmentation}

Translation augmentation allows a model trained in one language (source
language) to gain exposure to diverse linguistic contexts and structures
present in other languages. We expanded our English training dataset
by incorporating translations of training examples into our target
languages. For obtaining translations, we used the Opus-MT\footnote{\url{https://github.com/Helsinki-NLP/Opus-MT}} model from Helsinki-NLP, accessible via the Easy-NMT\footnote{\url{https://github.com/UKPLab/EasyNMT}} package, which we chose as a competitive and open source solution.
By providing translations of the training data into multiple
languages, the model can learn to recognize common patterns, sentiment
expressions, and stance cues that transcend linguistic boundaries.
Different languages may employ distinct grammatical structures, vocabulary,
and cultural nuances. By training on a multilingual dataset that includes
augmented training data incorporating translations, the model learns to identify semantically equivalent expressions and sentiment cues in different languages. 

\subsection{Multilingual Encoder and Stance classifier}

Multilingual BERT (mBERT), pre-trained on 104 languages, has been
shown to perform well for zero-shot cross-lingual transfer, especially
when fine-tuned on downstream tasks. Therefore, we employed mBERT
to finetune a multilingual encoder that can generate contextual embeddings
tailored to our task-specific tweets. Simultaneously, we jointly trained
a stance classifier, optimized to categorize stances based on these
embeddings. 

\subsection{Adversarial Language Adaptation}

Our approach for language adaptation was inspired by the adversarial
domain adaptation (ADA) with distillation technique proposed by Ryu
and Lee (2020) for unsupervised domain adaptation \cite{Ryu_Lee_2020_AAD}.
Their method, ADA, aims to enhance the performance of BERT in the
face of domain shifts. ADA combines the adversarial discriminative
domain adaptation (ADDA) framework \cite{Tzeng_etal_2017_ADDA} with
knowledge distillation. 

To adapt our model to the target languages, we utilize unlabeled French,
German, and English tweets from the VaccinEU dataset. Our adaptation
process comprises three key steps, described as follows:

\begin{enumerate}
 \item In Step 1, the multilingual encoder and stance classifier are fine-tuned
on labeled stance data. 

 \item In Step 2, the target language encoder is adapted through adversarial
training and distillation. This involves training the target encoder
and discriminator alternately in a two-player game. The discriminator
learns to distinguish between English and non-English representations,
while the target encoder learns to confound the discriminator. Knowledge
distillation serves as a regularization technique, preserving information
learned from the English training data. 

 \item Finally, in Step 3, the adapted target encoder and stance classifier
are utilized to predict stance labels for data in the target languages. 
\end{enumerate}

We release the code used to train and evaluate our models on GitHub.\footnote{\url{https://github.com/Bharathi-A-7/MTAB-cross-lingual-vaccine-stance-detection}}

\section{Experiments}
\label{sec:experiments}

In our experiments, we evaluate three fundamental approaches using
multilingual BERT, both as standalone models and in combination with
language adversarial learning. Initially, we utilize the labeled training
data to train a stance classifier and a task-specific multilingual
encoder. Subsequently, we adapt the encoder to the target languages
French, German, and Italian by leveraging unlabeled data from the
VaccinEU dataset. The Generator used for our adversarial learning
experiments is the encoder for the target languages. Our primary focus
in these experiments is to assess the effectiveness of our proposed
MTAB that leverages adversarial learning and translation augmentation
and to see its impact on the performance of zero-shot cross-lingual
models. 

\subsection{Models}

We compare a range of models that we test against our proposed MTAB model. In the absence of existing models for zero-shot cross-lingual stance detection, we use a strong, general-purpose baseline model, Multilingual BERT (mBERT). In addition, we also compare with an ablated version of our MTAB model, which we refer to as MTAB-noTL. We describe next the models we compare in our experiments:

\begin{itemize}
 \item \textbf{mBERT - Baseline model:} This model involves fine-tuning a multilingual
instance of the BertForSequenceClassification model from HuggingFace
on English training data. It serves as our baseline model for stance
detection. 

 \item \textbf{MTAB:} This is our model proposed above in Section \ref{sec:mtab}. It contains
the combined English vaccine stance dataset along with translations
of the English training set into French, German, and Italian. 

 \item \textbf{MTAB-noTL:} This is an ablated version of our proposed MTAB,
which does not make use of translation-augmented data and therefore
could be possibly limited for the capabilities of the adversarial
learning component. This model in turns tests the contribution of our proposed adversarial learning component on top of the translated data.
\end{itemize}

For each of the three models above, we test two different architectures, with and without the adversarial learning component, which we indicate in the name of the model by either including or not `+ Adversarial learning' at the end of it. This leaves us with a total of six model variants, i.e. the three models listed above with and without adversarial learning component.

\subsection{Model Hyperparameters}

We experiment with various sets of hyperparameters for our models on held-out data,
and selected the ones that yielded the best results. Across all our
experiments, we maintained a maximum sequence length of 128 and a
batch size of 32. However, the learning rate and the number of epochs
varied among the three different architectures. 

In the case of MTAB, we conducted training for 9 epochs using
an Adam Optimizer. The learning rate was set at $5e^{-5}$, facilitating
the joint training of the language encoder and stance classifier.
For the adversarial learning setup in MTAB, again we used an
AdamW Optimizer. The learning rate was $1e^{-5}$ for both the
Discriminator and the Generator, and each of the target languages
went through 5 training epochs.

\begin{table*}
\begin{centering}
\begin{tabular}{|c|c|c|c||c|}

\hline 
\multirow{1}{*}{Model} & VaccinEU - French & VaccinEU - German & VaccinEU - Italian & Average \tabularnewline
\hline 
mBERT & 0.50 & 0.50 & 0.46 & 0.487 \tabularnewline
\cline{2-5}
mBERT + Adversarial learning & 0.16 & 0.51 & 0.38 & 0.350 \tabularnewline
\hline 
MTAB-noTL & 0.45 & 0.51 & 0.39 & 0.450 \tabularnewline
\cline{2-5}
MTAB-noTL + Adversarial learning & 0.51 & 0.55 & 0.47 & 0.510 \tabularnewline
\hline 
MTAB & 0.51 & 0.53 & 0.45 & 0.497 \tabularnewline
\cline{2-5}
MTAB + Adversarial learning & \textbf{0.52} & \textbf{0.56} & \textbf{0.48} & \textbf{0.520} \tabularnewline
\hline 
\end{tabular}\caption{Res\label{tab:Results-of-Experiments}ults of Experiments (F1 scores).}
\par\end{centering}
\end{table*}

\section{Analysis of Results}
\label{sec:results}

Table \ref{tab:Results-of-Experiments} presents the results of our
experiments using F1 scores. A detailed examination of the results
offers various insights into the effects of model architecture on
the language embeddings and subsequently the classifier performance.
The combination of MTAB and adversarial language adaptation
in the presence of translation augmented data emerges as the top-performing
model, showcasing exceptional performance across the three target
languages. This underscores the effectiveness of translation augmentation
coupled with adversarial learning in achieving robust cross-lingual
transfer. Interestingly, the MTAB model leveraging adversarial learning outperforms all other models consistently across the three target languages, and consequently also on average across all three languages; our full MTAB model with adversarial learning (0.520) outperforms the ablated variant with no translations by an absolute 1\% (0.510) and the ablated variant with no adversarial training by 2.3\% (0.497). Having looked at the overall performance scores, we next delve into different aspects of the model to better understand how this improvement happens.

\textbf{\textit{Impact of Model Architecture:}}\textbf{ }The decision
to split the mBERT baseline model into separate models for language
encoding and classifier models, as implemented in MTAB, yields
notable benefits. This separation allows for higher control and flexibility
over the model's parameters. Our experiments and analysis
reveal that the adversarial training in mBERT + Adversarial Learning,
which simultaneously updates both the language encoder and classifier
parameters, is very unstable and adversely impacts performance, especially
for the French test set. In contrast, MTAB + Adversarial Learning,
which only updates the parameters of the target language encoders
while keeping the classifier parameters fixed, demonstrates superior
performance.

\begin{figure*}
\begin{centering}
\subfloat[French Test set.]{\includegraphics[width=3.5in]{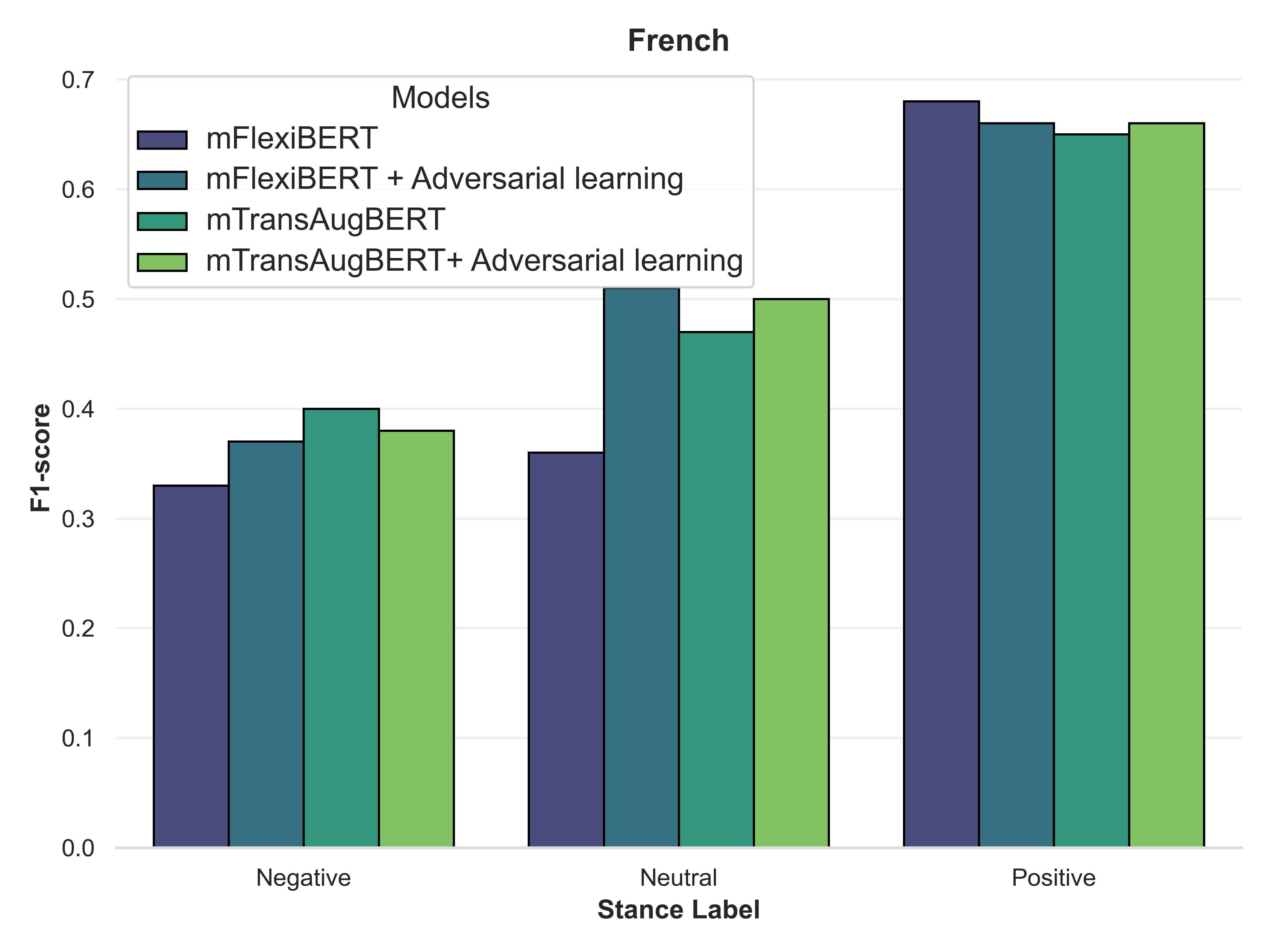}

}\subfloat[German Test set.]{\includegraphics[width=3.5in]{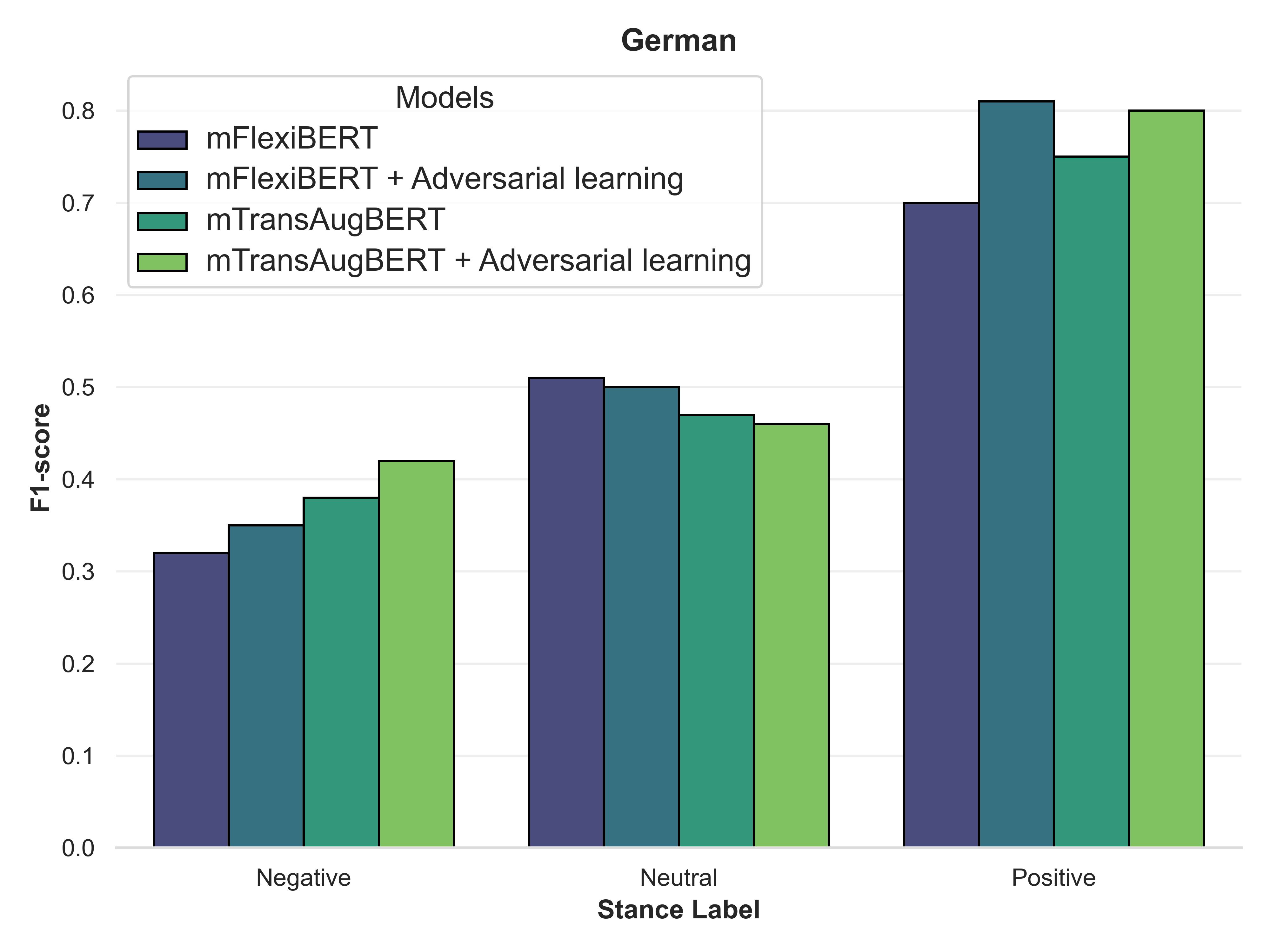}

}
\par\end{centering}
\begin{centering}
\subfloat[Italian Test set.]{\includegraphics[width=3.5in]{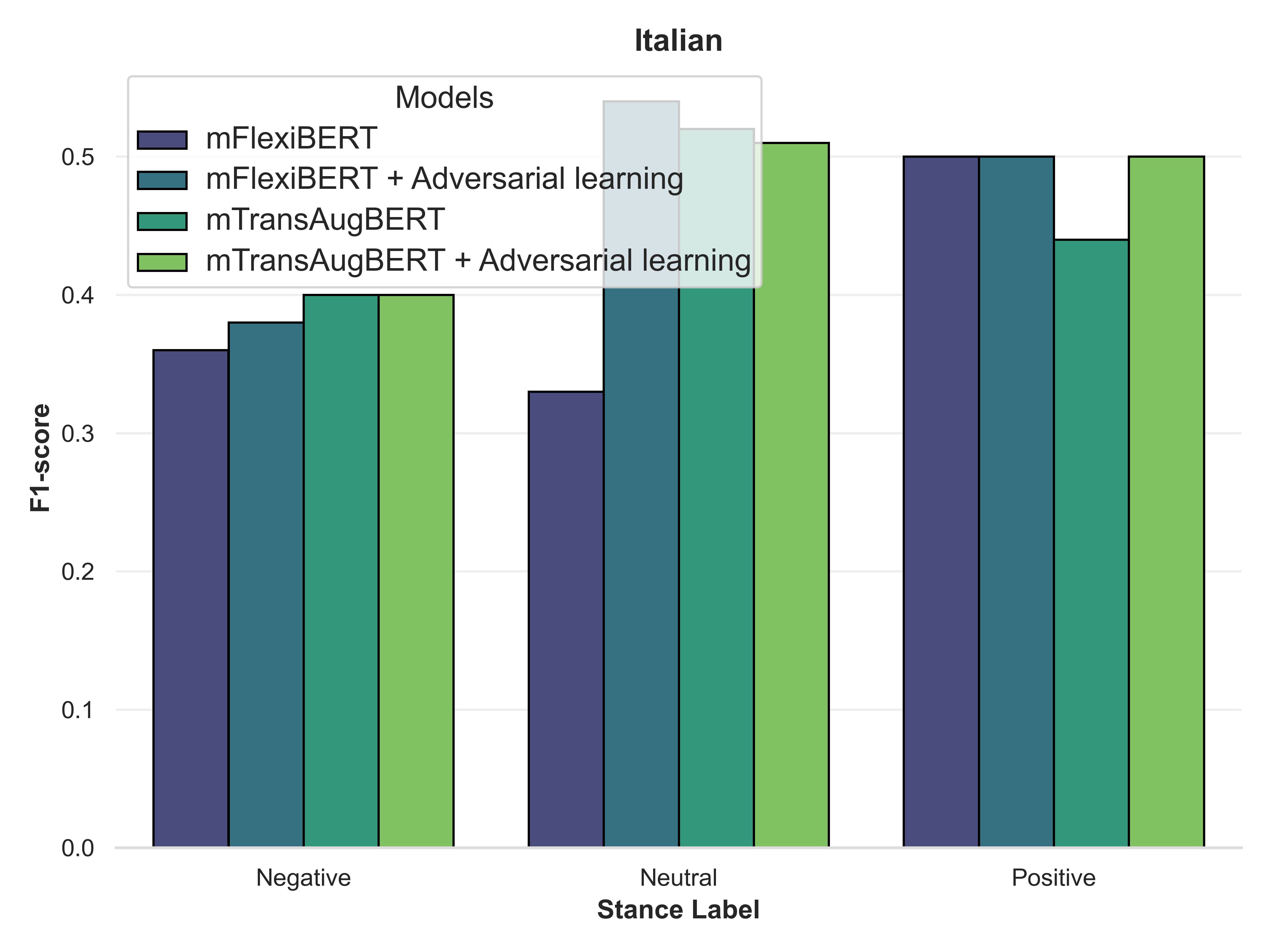}

}
\par\end{centering}
\caption{\label{fig:Class-wise-F1-scores}Breakdown of the MTAB model performance showing class-wise F1-scores.}

\end{figure*}

\textbf{\textit{Effectiveness of Adversarial learning:}} In both MTAB-noTL
and MTAB, the addition of an adversarial language adaptation
component contributes significantly to the F1 scores across the three
target languages in the test set. As demonstrated in several other
research works, this highlights the positive impact of adversarial
learning on cross-lingual generalization. As stated above, however, the contribution of the adversarial learning component is notable when this is carefully integrated into the MTAB model, but fails to contribute to the baseline MBERT model.

\begin{figure*}
\begin{centering}
\subfloat[French]{\includegraphics[width=0.5\textwidth]{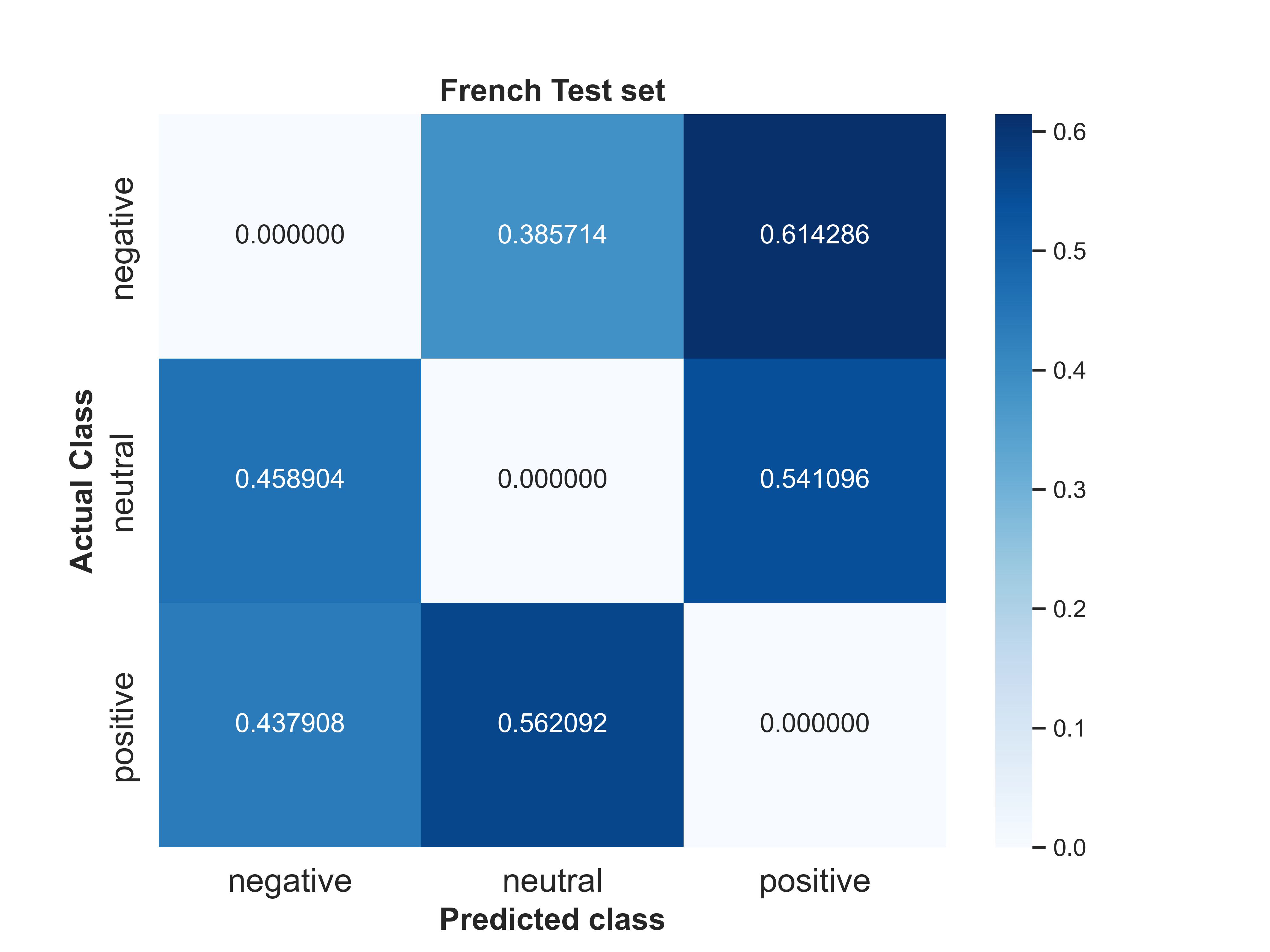}

}\subfloat[German]{\includegraphics[width=0.5\textwidth]{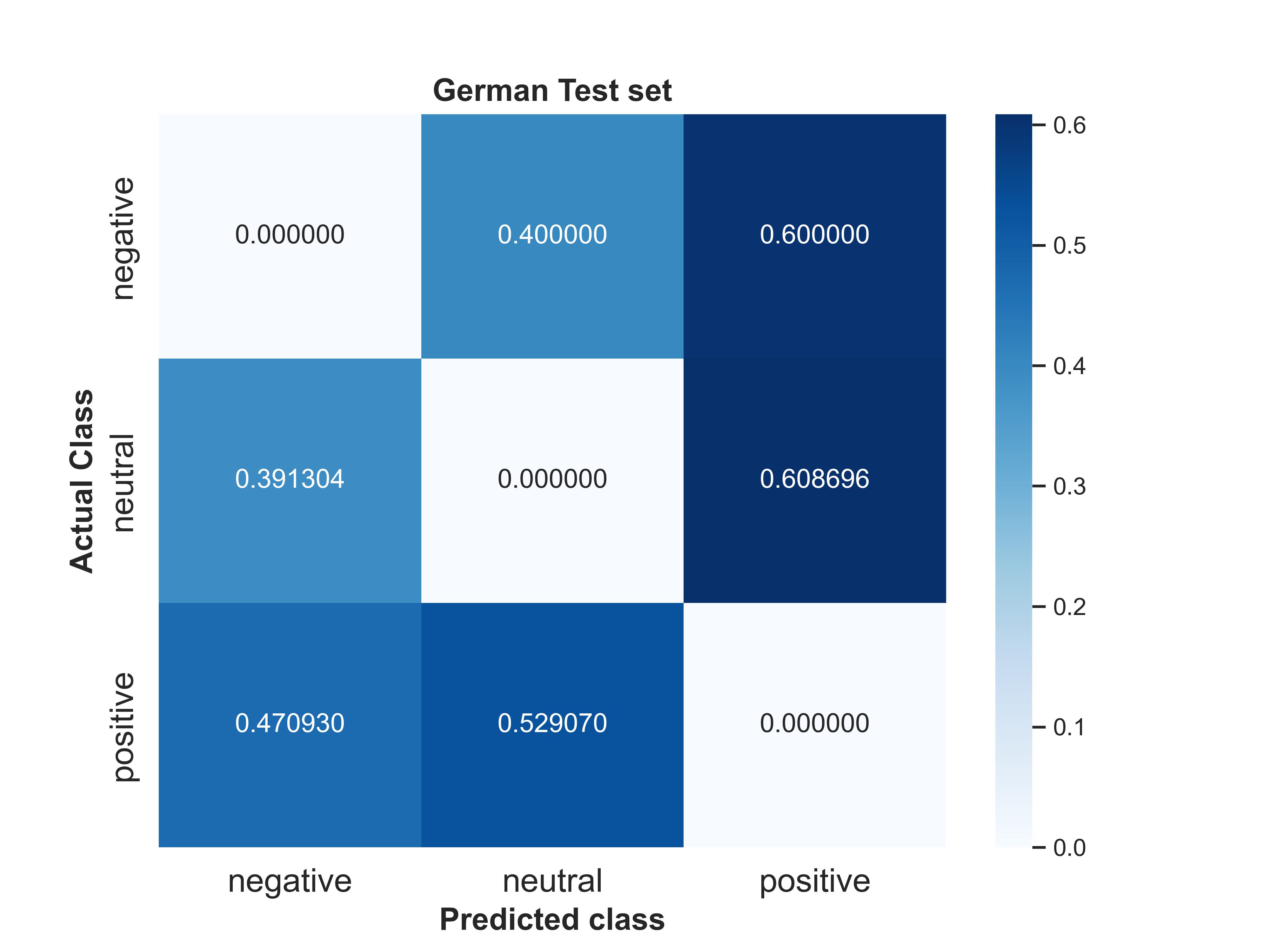}

}
\par\end{centering}
\begin{centering}
\subfloat[Italian]{\includegraphics[width=0.5\textwidth]{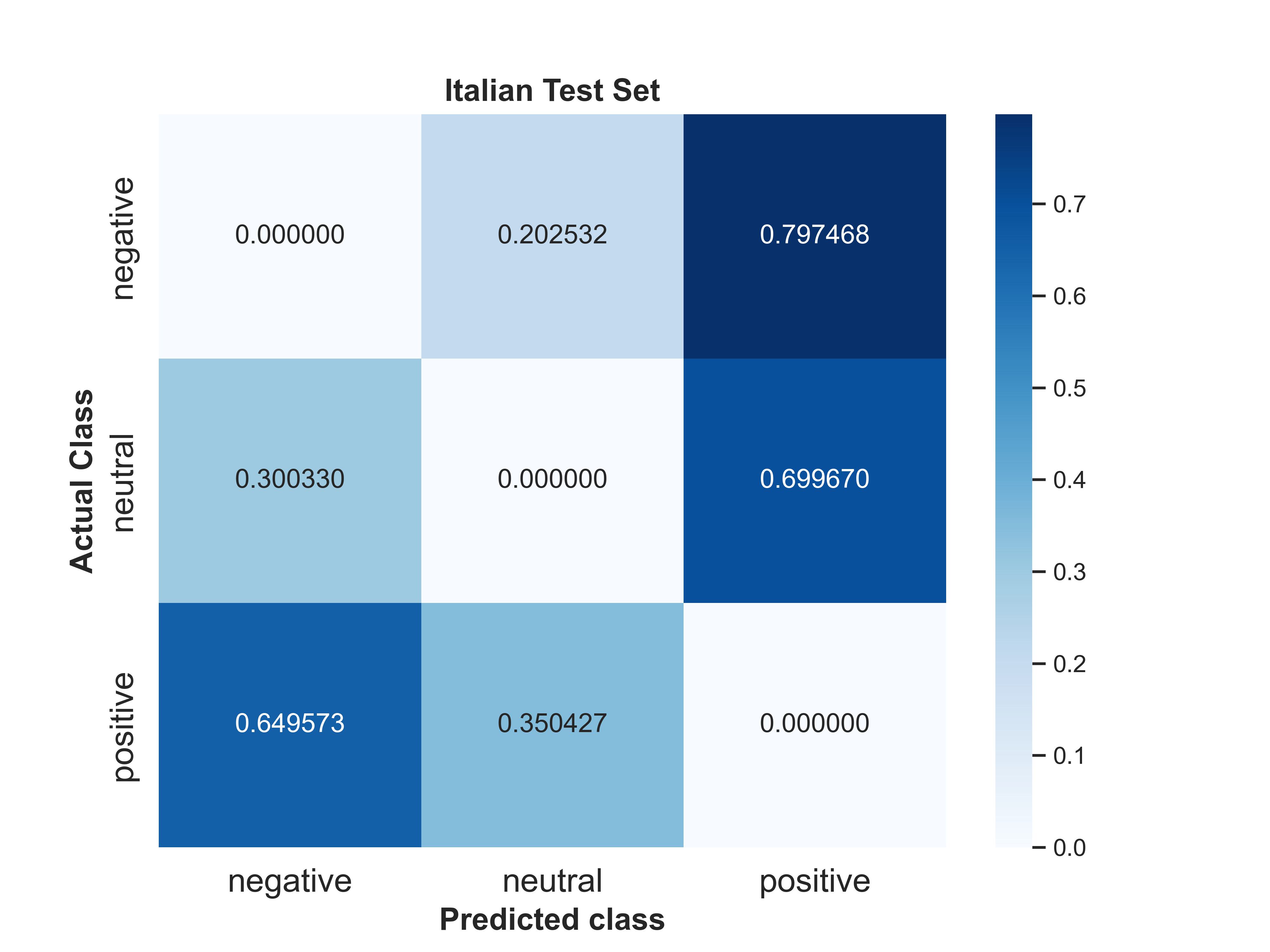}

}
\par\end{centering}
\caption{\label{fig:Distribution-of-Incorrect}Confusion matrix showing only the incorrect predictions made by the MTAB model.}

\end{figure*}

\textbf{\textit{Simple Efficacy of Translation Augmentation:}} The
simple and straightforward technique of augmenting training data with
translations of the English tweets, as employed in MTAB,
proves remarkably effective in enhancing the test scores in the absence
of labelled data for the target languages. This addition surpasses
the performance of MTAB-noTL both with and without an adversarial
learning component. 

The findings suggest that while adversarial learning enhances cross-lingual
generalization, translation augmentation further amplifies this effect.
This combination shows potential for handling cross-lingual tasks,
especially in low-resource languages with no labeled examples. 

However, there is a considerable imbalance among the data classes
as shown in Tables \ref{tab:Training-data} and \ref{tab:Test-data}. This imbalance is reflected
in the performance of the model on each of those respective classes.
Figure \ref{fig:Class-wise-F1-scores} shows the F1-scores of the
models across the three stance labels: Positive, Negative and Neutral.
There is a clear imbalance in the F1-scores with a difference of 10-20\%
between the minority classes Negative and Neutral and the majority
class Positive, which was expected given the dominance of the positive class in the data. This difference is more pronounced particularly for
the German language with the F1-scores for the positive label touching
as high as 80\% and for minority classes around 50\%. This overfitting
can also be understood from the Distribution of incorrect predictions
shown in Figure \ref{fig:Distribution-of-Incorrect}. The figures
show a consistent pattern where in more than 50\% of the time, the
Negative and Neutral classes are incorrectly classified as Positive.

While our research and proposed model MTAB contributed with the first ever zero-shot cross-lingual stance detection model, showing performance improvements over competitive baselines, we do believe that future research should further look into mitigating the impact of class imbalances, not least when these differences vary across languages.

\section{Conclusion}
\label{sec:conclusion}

Through the introduction of the novel model called Multilingual Translation-Augmented BERT (MTAB), our research represents the first effort in zero-shot
cross-lingual stance detection, where no labeled data is available during training for the target language.
Our proposed model MTAB makes the most of labeled
data available from a resource-rich language, such as English, to enable the
development of stance detection systems for less-resourced languages
with very minimal manual effort. In doing so, we also contribute to
the progression of multilingual vaccine stance detection methodologies.
Through experimentation with a range of datasets labeled for stance around vaccine hesitancy and support, we have explored the possibility of transferring knowledge from a resource-rich language such as English, to other languages for which fewer and smaller labeled datasets are available, which we tested with French, German and Italian. Comparing our model with a competitive baseline model, mBERT, as well as ablated variants of our model, we demonstrate the potential of MTAB to improve cross-lingual transfer for stance detection in zero-shot settings. Further delving into the contributions of different components of the model, we also observe the importance of leveraging translation-augmented data, as well as using the adversarial learning component, for improved performance, which all in all demonstrates the effectiveness of integrating all the components of MTAB.

While our primary focus remains on leveraging deep learning approaches,
we believe that future work in the area of online stance detection
can benefit greatly from recent progress in the development of Large
Language Models (LLMs). We also aim to overcome the limitations of an imbalanced
training dataset by exploiting the few-shot learning capabilities
of LLMs enabled via Prompt Engineering. The development of a fine-tuned
language model can also help understand the subtle nuances present
in the languages used within online social media pertaining to vaccines
and discussions related to public health.

\bibliographystyle{IEEEtran}
\bibliography{references}


 




\vfill

\end{document}